\documentclass[11pt,a4paper]{article}
\usepackage[hyperref]{acl2021}
\usepackage{times}
\usepackage{latexsym}
\usepackage{multirow}

\usepackage{graphicx}
\graphicspath{ {./images/} }

\usepackage{microtype}

\aclfinalcopy 
\def\aclpaperid{4} 

\title{Towards Equal Gender Representation in the Annotations of Toxic Language Detection}

\author{Elizabeth Excell \\
  Department of Computer Science \\
  Durham University \\
  Durham, UK \\
  \texttt{bethexcell@gmail.com} \\\And
  Noura Al Moubayed \\
  Department of Computer Science \\
  Durham University \\
  Durham, UK \\
  \texttt{noura.al-moubayed@durham.ac.uk} \\}

\date{June 2021}

\begin{document}
\maketitle
\begin{abstract}
Classifiers tend to propagate biases present in the data on which they are trained. Hence, it is important to understand how the demographic identities of the annotators of comments affect the fairness of the resulting model. In this paper, we focus on the differences in the ways men and women annotate comments for toxicity, investigating how these differences result in models that amplify the opinions of male annotators. We find that the BERT model associates toxic comments containing offensive words with male annotators, causing the model to predict 67.7\% of toxic comments as having been annotated by men. We show that this disparity between gender predictions can be mitigated by removing offensive words and highly toxic comments from the training data. We then apply the learned associations between gender and language to toxic language classifiers, finding that models trained exclusively on female-annotated data perform 1.8\% better than those trained solely on male-annotated data, and that training models on data after removing all offensive words reduces bias in the model by 55.5\% while increasing the sensitivity by 0.4\%\footnote{Paper is accepted at GeBNLP2021 workshop at  ACL-IJCNLP 2021}.
\end{abstract}

\section{Introduction}

Toxic language detection has attracted significant research interest in recent years as the volume of toxic user-generated online content has grown with the expansion of the Internet and social media networks \citep{schmidt2017survey}. As toxicity is such a subjective measure, its definition can vary significantly between different domains and annotators, leading to many contrasting approaches to toxicity detection such as evaluating the constructiveness of comments \citep{kolhatkar2020classifying} or examining the benefits of taking into account the context of comments \citep{pavlopoulos-etal-2020-toxicity}.

Detecting and appropriately moderating toxic comments has become crucial to online platforms to keep people engaged in healthy conversations rather than letting hateful comments drive people away from discussions. In addition, it has become increasingly important to ensure a user's right to free speech and only remove comments that violate the policies of the platform. Human annotators are the most effective way to filter toxic comments. However, they are costly and unscalable to the generated data. As such, toxic language classifiers are trained on datasets composed of comments annotated by humans as an efficient way of detecting toxic language \citep{schmidt2017survey}.

One of the main issues with this approach is that any biases held by the pool of annotators are propagated in the classifier, which can lead to non-toxic comments from certain identity groups being mislabelled as toxic, an effect known as false positive bias (\citealp{dixon2018measuring}; \citealp{sap-etal-2019-risk}). While many papers have acknowledged the potential for bias in their datasets, with some proposing novel ways of measuring this bias \citep{dixon2018measuring}, very little has been done to examine the differences in the ways that distinct groups of annotators perceive comments and investigate how these differences affect the classification results.

This paper is motivated by the lack of understanding of the impact of annotator demographics on bias in toxic language detection. We investigate how the annotators' demographics affect the toxicity scores/labels and the trained models. We analyse the chosen corpus by grouping the annotations by the gender of the annotator as it is the most addressed demographic variable in the literature and constitutes the largest groups of data in the corpus. We then tailor the state-of-the-art BERT model to the tasks of toxicity and gender classification, using training and test sets built independently using the annotations of different genders to investigate bias.

For the gender classification models, we use explainable machine learning methods to analyse the comments in the test set in order to gain further insights into the associations between gender and language made by the model that contribute to the biased classifications towards male annotators. We then explore how modifying the training data of the models based on these learned associations affects the bias present. We examine the role offensive language plays in male and female annotations and investigate the robustness of models trained independently on gender-specific data once offensive language has been removed.

The main contributions of this work are: I) revealing the bias of BERT-based toxic language detection models towards male annotators, II) recognising the learned associations between male annotators and offensive language in the model, III) demonstrating methods to reduce the bias in the model without reducing the sensitivity. 

\section{Bias Statement}

In this work, we explore gender bias present in toxic language detection systems due to associations between offensive language and annotator gender amplified by the model. We define gender bias in this context as the disproportionate influence of the opinions of one gender over another in the model's output. We acknowledge that by treating gender as binary in this study, we exclude those who identify as non-binary, which may cause representational harm \citep{blodgett-etal-2020-language}. This choice was made due to the scarcity of annotators who identify as non-binary affecting the generalisability of the results.

This work demonstrates that toxic comments containing offensive words are associated with male annotators, resulting in female annotators predicted as being male. This leads to toxicity classifiers that are overly reliant on the opinions of annotators perceived to be male in order to make a classification. The resulting systems create representational harm by overlooking the diverse opinions of female annotators, leading to comments that women may consider toxic not being removed.

\section{Related Work}

Previous research into gender bias in toxic language detection caused by the demographic makeup of annotators explored superficial differences between male and female annotators, but only reflected on the ethical considerations involved rather than thoroughly investigating the differences between annotator groups and attempting to minimise bias in the model. 

\citet{binns2017like} presented different methods for detecting potential bias by building classifiers trained on comments whose annotators belong to different genders. They reported differences in average toxicity scores and inter-annotator agreement between the groups. Similar work by \citet{sap-etal-2019-risk} in the field of racial bias examined toxicity scores given to Twitter corpora, where the white annotators in the majority give higher toxicity scores to tweets exhibiting an African American English dialect, demonstrating how annotator opinions can propagate bias throughout the model.

Some studies focused on gender bias in specific tasks in Natural Language Processing such as coreference resolution. The aim of those studies is to eliminate under-representation bias by applying gender-swapping and name anonymisation to a corpus to balance the use of gender-specific words \citep{zhao-etal-2018-gender}. \citet{sun-etal-2019-mitigating} highlights this technique as an effective way of debiasing models and measuring gender bias in predictions, using the False Positive Equality Distance (FPED) and False Negative Equality Distance (FNED) metrics \citep{dixon2018measuring} to measure the difference in performance for gender-swapped sentences.

Another common source of bias is the word embeddings, which can form associations between identity groups and stereotypical terms based on their prevalence in the literature used to train the language model. \citet{bolukbasi2016man} demonstrated the presence of gender bias in occupations in the word embeddings of a language model and proposed a system to debias those models by isolating the gender subspace before utilising hard or soft debiasing to remove the gender bias from terms identified as being gender neutral. This was further extended by \citet{manzini-etal-2019-black} to encompass racial bias, transforming the binary classification task of identifying gender-specific and gender neutral terms into a multiclass debiasing problem.

Related studies into the aggregation of crowdworker annotations highlight that many models are skewed towards the opinions of workers who agree with the majority vote, which can lead to the opinions of other annotators being disregarded even when there is low inter-annotator agreement \citep{balayn2018characterising}. A solution to this, proposed by \citet{aroyo2013crowd} and adopted by \citet{wulczyn2017ex}, uses disaggregated data and transforms the problem from the binary classification of toxicity to the prediction of the proportion of annotators who would classify a comment as toxic.

In practice, the effectiveness of crowdsourcing appears to be mixed for much of the literature, with \citet{kolhatkar2020classifying} noting that expert annotators only agree with the majority opinion of the crowdsourced annotations 87\% of the time in the context of evaluating the constructiveness of comments. This  verdict is also reached by \citet{nobata2016abusive}, who concludes that workers on the Amazon Mechanical Turk platform exhibit a much worse inter-annotator agreement than the in-house annotators for the task of abuse classification. This highlights the need to thoroughly examine the annotations in corpora before they are applied to a classification task.

We note that that the majority of the research into bias in toxic language detection does not reflect on the bias caused by the pool of annotators, and yet research into crowdsourcing demonstrates poor inter-annotator agreement in many corpora and how the results of classification models are skewed by annotator opinions that may not reflect society as a whole. For the few papers that do examine the role of annotators in toxic language detection, no practical suggestions have been made that aim to reduce the identified bias in the implemented model, which is the main contribution of this paper.

\section{Data}

We use the toxicity corpus\footnote{\url{https://figshare.com/articles/dataset/Wikipedia\_Talk\_Labels\_Toxicity/4563973}} from the Wikipedia Detox project \citep{wulczyn2017ex}, which contains over 160k comments from English Wikipedia annotated with toxicity scores and the demographic information of the annotators, where each comment has been labelled by approximately 10 annotators using the toxicity categories displayed in Table~\ref{question-table}.

This corpus has been widely used in recent literature developing deep learning approaches to toxic language detection (\citealp{pavlopoulos-etal-2017-deep}; \citealp{mishra-etal-2018-neural}) and investigating bias, such as \citet{dixon2018measuring} using the comments to propose metrics that evaluate bias based on the identity terms present in the data. As such, this corpus was selected for the comparability of results it provides, in addition to it being the only toxic language corpus to provide the genders of the annotators.

\citet{binns2017like} demonstrates methods to explore potential bias in this corpus without further investigating the cause of the bias or attempting to reduce bias in the model, finding that male annotators in the corpus have a significantly higher inter-annotator agreement than female annotators, leading to male test data performing better than female test data. \citet{balayn2018characterising} uses this corpus to investigate how the implemented model became skewed towards the scoring of annotators with the majority opinion, favouring the opinion of the largest group for each demographic variable. \citet{balayn2018characterising} then attempts to mitigate this bias by balancing the dataset for each demographic variable, which we discover is not enough to prevent bias is the model due to the learned associations between the demographic variable and the language in the comments. 

We hypothesise based on previous research that models trained on this corpus will likely value the opinions of male annotators over female annotators. This is due to the fact that male annotators were found to have a greater inter-annotator agreement than female annotators, meaning that they are likely to hold the majority opinion, and so it follows that the model will place a greater importance on the scores of male annotators when deciding the toxicity of a comment.

\begin{table*}
\centering
\resizebox{\linewidth}{!}{%
\begin{tabular}{lcl}
\hline \textbf{Toxicity Category} & \textbf{Toxicity Score} & \textbf{Description}\\ \hline
Very toxic & -2 & A very hateful, aggressive, or disrespectful comment\\
 & & that is very likely to make you leave a discussion \\
Toxic & -1 & A rude, disrespectful, or unreasonable comment that is\\
 & & somewhat likely to make you leave a discussion\\
Neither & 0 & - \\
Healthy contribution & 1 & A reasonable, civil, or polite contribution that is\\
 & & somewhat likely to make you want to continue a discussion \\
Very healthy contribution & 2 & A very polite, thoughtful, or helpful contribution that is\\
 & & very likely to make you want to continue a discussion \\
\hline
\end{tabular}}
\caption{\label{question-table} Toxicity categories given to annotators with associated toxicity scores and descriptions.}
\end{table*}

\section{Experiments}

\subsection{Technical Specifications}

We use a state of the art model \citep{zorian2019debiasing}, built based on the pre-trained uncased BERT\textsubscript{BASE} model \citep{devlin-etal-2019-bert} with a single linear classification layer on top. The Huggingface {\fontfamily{qcr}\selectfont transformers} library \citep{wolf-etal-2020-transformers} is used to implement the model. 

For fine-tuning, we follow the guidelines set by \citet{devlin-etal-2019-bert}, using an Adam optimizer with a learning rate of $2 \times 10^{-5}$ and a linear scheduler. We use a batch size of 8 trained over 2 epochs \footnote{Code is available at: \url{https://github.com/MicrosoftExcell/Advanced-Project}}. 

\subsection{Preliminary Data Analysis}

Examining the chosen corpus, we find that 34\% of the annotations were made by women (with $<$0.1\% of annotators describing themselves as `other'). Due to the unbalanced nature of the dataset, we balance each training and test set used for gender classification by ensuring that 50\% of the annotations were made by men and 50\% of the annotations were made by women. We achieve this by randomly sampling the comments annotated by each demographic group until a quota such as the size of the smallest group is reached for each sample. The goal of this is to eliminate under-representation bias in order to be certain that any differences between genders in the results are not caused by an unbalanced dataset.

After reviewing the toxicity scores given by each group as a whole, we find that female annotators on average annotated 1.72\% more comments as toxic than male annotators and assigned toxicity scores that were on average 0.048 lower than those given by their male counterparts, using the toxicity scores given in Table~\ref{question-table}. These figures indicate a slight disparity between the genders, suggesting that female annotators on average find comments more toxic than male annotators.

\subsection{Pre-processing}

While the different models built for this paper focus on two different tasks, namely toxicity and gender classification, the pre-processing steps remain largely the same. Firstly, the data is stripped of unnecessary information such as newline and tab tokens. Annotators who reported their gender as `other' are removed as they do not provide a large enough group to draw generalisable conclusions from. The dataset is then balanced by gender as previously described as well as being balanced by the toxicity score in a similar manner. 

For gender classification, as only toxic data is used for training and testing, this means sampling the data evenly from comments given a toxicity score of -1 and those given a toxicity score of -2. This is necessary as far fewer comments are labelled as `Very Toxic' than `Toxic', and as it is the toxic data that is being investigated, it is important to ensure that any differences in the way men and women annotate comments as `Very Toxic' are not diminished in the results by the substantial size of the `Toxic' category. Similarly, the toxicity classification models take 25\% of their data from the comments annotated as `Toxic' and a further 25\% from the `Very Toxic' data, with the remaining 50\% being randomly sampled from the `Healthy' and `Very Healthy' data. The last two categories were not divided evenly as with the toxic categories due to the limited size of the `Very Healthy' data.

We choose the maximum sequence length for the model to be 100 based on the token counts of comments in the training data, taking into account memory restrictions.

\subsection{Gender Classification}

The results of the preliminary data analysis indicate potential differences between male and female annotators in the corpus. We explore this further by tasking the BERT-based model with classifying the gender of an annotator based on a comment the annotator labelled as toxic.

\begin{figure}
    \centering
    \includegraphics[scale=0.39]{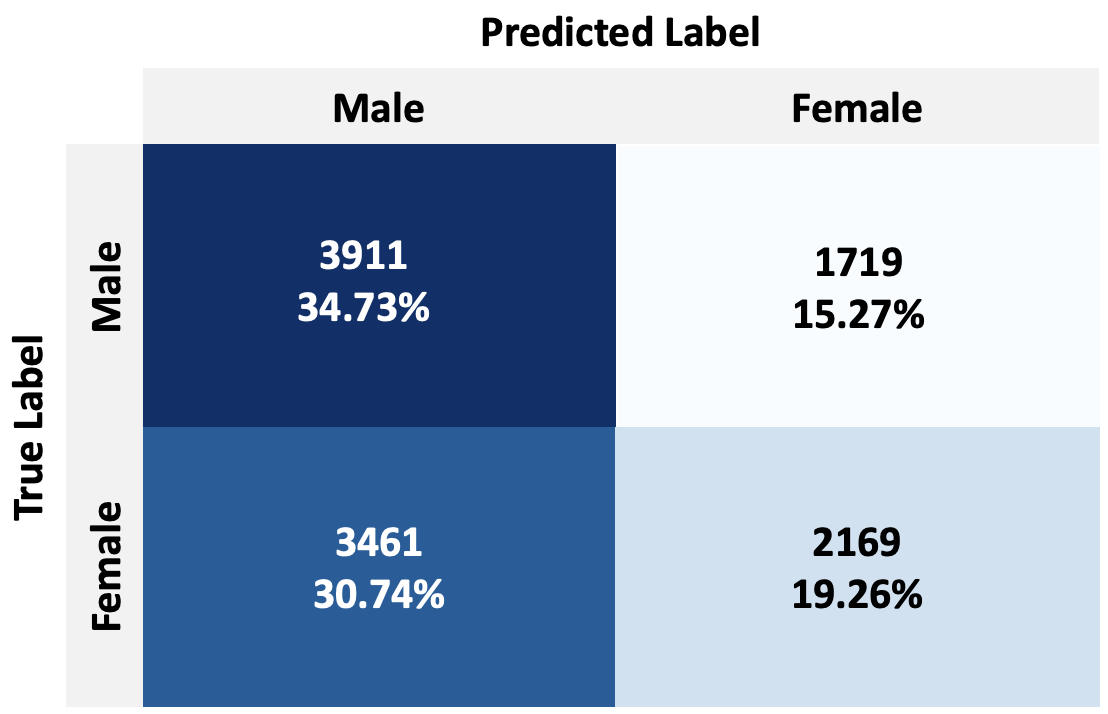}
    \caption{Confusion matrix showing the gender predictions of the annotators of toxic comments by the BERT-based model.}
    \label{fig:gendermatrix}
\end{figure}

Using training and test data classified as toxic or very toxic by equal numbers of male and female annotators, we find that the model predicts the gender of the annotator of a toxic comment as male 67.7\% of the time on average, with the results of the first run shown in Figure~\ref{fig:gendermatrix} . This indicates that there is a difference between the annotations of male and female annotators that can be identified by the model, as we would expect the predictions to be evenly distributed between male and female if no bias was present. 

In order to investigate the differences in annotation styles between the genders that caused the bias shown, we add interpretability to the model's output by adapting the attribution scores and integrated gradients to display which words in comments are the most important when predicting the gender of the annotator, and which gender those words are attributed to. The integrated gradients method attributes the predictions of deep networks to their inputs and has proven useful for rule extraction in text models, identifying undiscovered correlations between terms and classification results \citep{sundararajan2017axiomatic}.

\begin{table*}
\centering
\resizebox{\linewidth}{!}{%
\begin{tabular}{lllrr}
\hline \textbf{True Label} & \textbf{Predicted Label} & \textbf{Attribution Label} & \textbf{Score} & \textbf{Word Importance}\\ \hline
female & male (0.53) & female & -0.70 & \raisebox{-1pt}{\includegraphics[scale=0.85]{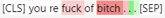}}\\
female & male (0.53) & female & -2.06 & \raisebox{-1pt}{\includegraphics[scale=0.85]{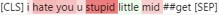}}\\
female & male (0.53) & female & -1.33 & \raisebox{-1pt}{\includegraphics[scale=0.85]{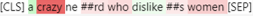}}\\
female & male (0.53) & female & -1.49 & \raisebox{-1pt}{\includegraphics[scale=0.85]{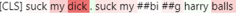}}...\\
male & female (0.54) & female & 2.00 & \raisebox{-1pt}{\includegraphics[scale=0.85]{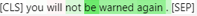}}\\
male & female (0.56) & female & 1.79 & \raisebox{-1pt}{\includegraphics[scale=0.85]{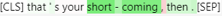}}\\
male & male (0.53) & female & -1.39 & \raisebox{-1pt}{\includegraphics[scale=0.85]{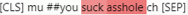}}\\
male & male (0.56) & female & -1.89 & \raisebox{-1pt}{\includegraphics[scale=0.85]{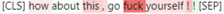}}\\
female & female (0.71) & female & 4.06 & ...\raisebox{-1pt}{\includegraphics[scale=0.85]{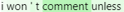}}...\\
 & & & & ...\raisebox{-1pt}{\includegraphics[scale=0.85]{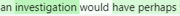}}...\\
female & female (0.67) & female & 5.00 & ...\raisebox{-1pt}{\includegraphics[scale=0.85]{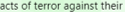}}...\\
 & & & & ...\raisebox{-1pt}{\includegraphics[scale=0.85]{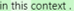}}...\\
\hline
\end{tabular}}
\caption{\label{interpret-table} Attributions of annotator gender to words in toxic comments. First column contains the true gender of the annotator. Second column contains the predicted gender of the annotator with the associated probability given by the BERT model. Third column contains the attribution label. Fourth column contains the attribution score, for comparison with the attribution label (negative scores indicate male attributions and positive scores indicate female attributions). Fifth column contains the comment text highlighted with the associated word attribution scores. \textbf{Red indicates negative (male) attribution scores, green indicates positive (female) attribution scores.} The intensity of the colour indicates the magnitude of the associated attribution. \textit{Note: some comments are truncated due to their length, in which case the words with the strongest attribution scores are shown.}}
\end{table*}

The results of this analysis can be seen in Table~\ref{interpret-table}, where 10 comments from the test set have been chosen due to their brevity and concise representation of the attribution scores seen in the test set as a whole. Furthermore, we include comments from each combination of true and predicted labels to provide a wider picture of the observed results.

We observe that the model gives great importance to offensive words when classifying a comment as having a male annotator. The language in comments predicted as having a female annotator is less explicit and harder to categorise, other than that the attributed words are more typical of a conversation rather than an overt insult like the majority of the male attributed words. This is corroborated by the Spearman's rank correlation coefficient of -0.378 between the probability given by the model of the annotator being female and the number of offensive words in the comment, indicating the existence of a relationship between the model predicting annotators as being male and the presence of offensive words in a comment.

Examining the data further, we find that male-annotated `Toxic' comments contains 0.1 more offensive words on average than female-annotated `Toxic' comments, with this disparity rising to 0.28 for the `Very Toxic' comments.

Based on these observations, we hypothesise that the bias of the model towards predicting a toxic comment as having a male annotator is due to the model learning an association between offensive words and male annotators in the training data, exacerbated by the prevalence of offensive words in toxic comments. In order to validate this hypothesis, we retrain the model after removing all offensive words from the training data using a blacklist\footnote{\url{https://www.cs.cmu.edu/~biglou/resources/}}. We refer to the original BERT model as BERTOriginal and this new model as BERTNoProfanity.

We also train the model after removing the `Very Toxic' data in addition to the offensive words, in order to see if this lessens the gender disparity in the results. We do this based on the knowledge that the most toxic comments contain the greatest amount of profanity as comments annotated as `Toxic' have a median of 1 and a mean of 1.20 offensive words per comment, while the `Very Toxic' comments have a median of 2 and a mean of 2.41 offensive words per comment. This new model is referred to as BERTNotVeryToxic.



\begin{figure}
    \centering
    \includegraphics[scale=0.4]{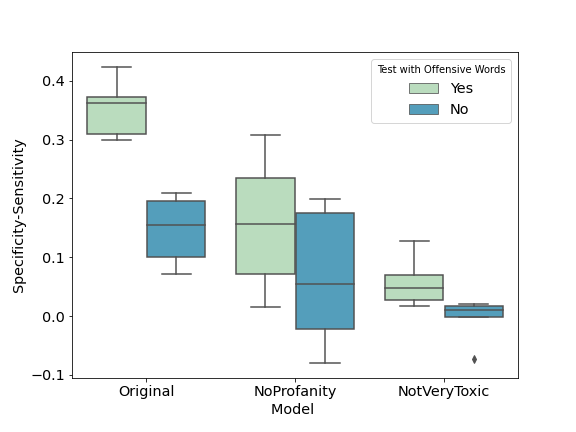}
    \caption{Box plot showing the difference between the specificity and sensitivity for each gender classification model on test data with and without offensive words.}
    \label{fig:diffplots}
\end{figure}

The performance of these models on toxic test data with and without offensive words is displayed in Figure~\ref{fig:diffplots}. We measure the difference between specificity and sensitivity for each model as they measure the model's ability to correctly predict whether an annotator is male or female respectively. Ideally, all values of specificity and sensitivity should be 0.5 if there is no bias towards either gender in the results. As such, the difference between them is indicative of the amount of bias in the model.

What we observe from these results is that bias is reduced in all models when offensive words are removed from the test data, indicating that the offensive words are a large contributor to the bias towards predicting annotators as male. We also note that the BERTNoProfanity model shows a 55.5\% reduction in bias on average compared to the BERTOriginal model, again demonstrating that offensive words cause bias in the model. Furthermore, we see that the BERTNoProfanity model exhibits the greatest amount of variation in the results, due to the discrepancies in the semantics between comments with and without words removed. The BERTNotVeryToxic model does not face this issue as it is trained using only the `Toxic' data, which has half the number of offensive words per comment than the `Very Toxic' data does, meaning that the semantics of comments remain broadly intact.

In addition, we observe that the BERTNotVeryToxic model exhibits the least bias overall, suggesting that the `Very Toxic' data contributes to the model's decision to predict the gender of an annotator as male. In fact, the BERTNotVeryToxic model exhibits little to no bias on the test data without offensive words, apart from one outlier that leans towards female predictions, suggesting that the bias towards men is eliminated when offensive words and the `Very Toxic' data are removed from the training and test data.

\begin{figure*}[h]
    \centering
    \includegraphics[scale=0.49]{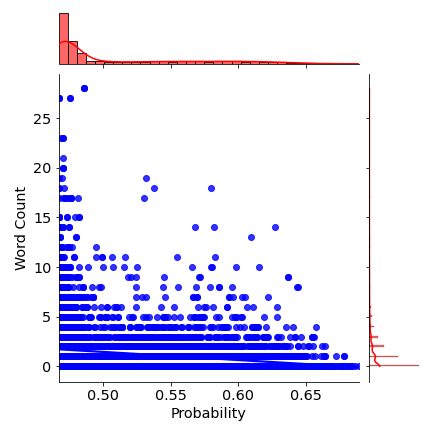} \hspace{1cm} \includegraphics[scale=0.48]{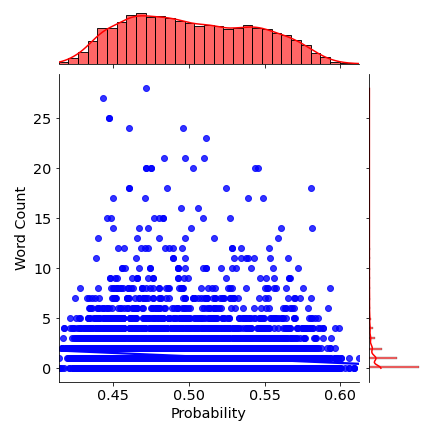}
    \caption{Scatter plots showing the number of offensive words in a comment against the predicted probability of the annotator being female for the BERTOriginal model (left) and the BERTNotVeryToxic model (right).}
    \label{fig:bertplots}
\end{figure*}

In order to further validate our hypothesis about the relationship between gender predictions and offensive words in comments, we plot the relationship between the predicted probability of a comment having a female annotator and the number of offensive words in the comment for the BERTOriginal and BERTNotVeryToxic models, the results of which can be seen in Figure~\ref{fig:bertplots}.

From these plots we can see that the BERTOriginal model is very likely to make gender predictions based on the number of offensive words in a comment as the probability distribution is skewed towards the left, meaning that comments with high numbers of offensive words have low probabilities of being female. We can see that this is not the case for the BERTNotVeryToxic model, as it shows a much more even distribution of gender probabilities for comments with higher numbers of offensive words, again confirming the model's reliance on `Very Toxic' data to make the association between male annotators and offensive words in toxic comments.

\begin{figure*}
    \centering
    \includegraphics[scale=0.39]{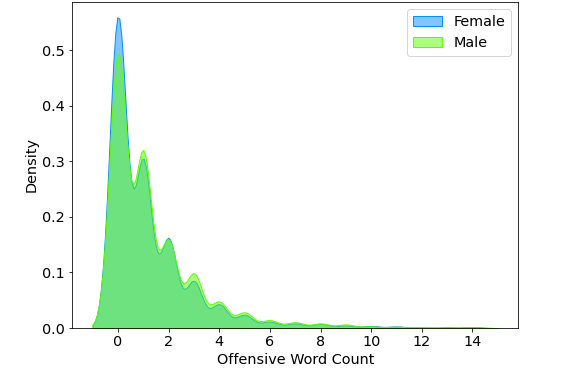} \includegraphics[scale=0.39]{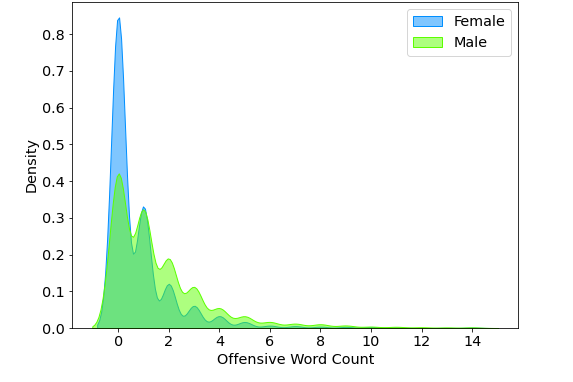}
    \caption{Kernel density estimation of the probability distribution of the count of offensive words in comments for the ground truth (left) and predicted (right) male and female labels.}
    \label{fig:genderlabels}
\end{figure*}

In order to demonstrate that the number of offensive words in a comment is not a reliable method of predicting the gender of an annotator, we examine the true and predicted labels of all comments in the test set, as can be seen in Figure~\ref{fig:genderlabels}. This shows that both men and women annotate comments with a high number of offensive words as toxic, as the estimation of the probability distribution for the true gender labels is roughly the same for both genders. We can see that this distribution has shifted in the predicted labels, with the female distribution being shifted to the left and the male distribution being shifted to the right. This shows that the model attributes comments with no offensive words to female annotators and comments with greater numbers of offensive words to male annotators despite there being little difference between the gender distributions in the ground truth.

\subsection{Toxicity Classification}

To further explore the differences between male and female annotators, we adapt the BERT model to perform toxicity classification rather than gender classification. For this task, we keep the dataset balanced between toxic and non-toxic comments. The model is trained using data from male and female annotators respectively, with and without offensive words removed. We refer to the male models with and without offensive words as BERTMale and BERTMaleNoProfanity respectively, and refer to the female models as BERTFemale and BERTFemaleNoProfanity in the same way.

We test each of the models using test data of the same condition as well as the test data from all other toxicity classification models. This means that models trained exclusively on data from one gender can be compared using data from both genders to examine which model performs better in addition to finding which set of test data is easier to categorise. This also allows us to examine the performance of models trained and tested on data with and without offensive words in order to understand the impact of removing offensive words from the training data on performance, as we have already determined that this method decreases bias in the model.

As we have only examined the relationship between annotator gender and the language in comments that were annotated as toxic, we use sensitivity to measure the performance of each model and set of test data. This measures the ability of each model to correctly classify toxic comments.

\begin{table*}[h]
\centering
\resizebox{\linewidth}{!}{%
\begin{tabular}{l|cc|cc|cc|cc|}
\cline{2-9}
  \multirow{2}{*}{} & \multicolumn{8}{|c|}{\textbf{Test Data}} \\
\cline{2-9}
  & \multicolumn{2}{|c}{Male} & \multicolumn{2}{|c}{MaleNoProfanity} & \multicolumn{2}{|c}{Female} & \multicolumn{2}{|c|}{FemaleNoProfanity}\\ 
\hline \multicolumn{1}{|l|}{\textbf{Model}} & Mean & SD & Mean & SD & Mean & SD & Mean & SD\\
\hline 
\multicolumn{1}{|l|}{BERTMale}  & 0.8370 & 0.019 & 0.6794 & 0.026 & 0.7682 & 0.019 & 0.6288 & 0.025\\
\multicolumn{1}{|l|}{BERTMaleNoProfanity} & 0.8392 & 0.004 & 0.8142 & 0.003 & 0.7748 & 0.008 & 0.7502 & 0.007\\
\multicolumn{1}{|l|}{BERTFemale} & \textbf{0.8534} & 0.004 & 0.6952 & 0.016 & \textbf{0.7986} & 0.013 & 0.6500 & 0.020\\
\multicolumn{1}{|l|}{BERTFemaleNoProfanity} & 0.8528 & 0.006 & \textbf{0.8224} & 0.007 & 0.7944 & 0.010 & \textbf{0.7662} & 0.012\\
\multicolumn{1}{|l|}{BERTMale+Female} & 0.8519 & - & 0.7376 & - & 0.7689 & - & 0.6682 & - \\
\hline
\end{tabular}}
\caption{\label{toxicity-sensitivity-table} Means and standard deviations of sensitivity for each toxicity model on each category of test data over 5 runs using seed values 42, 5936, 9743, 14280, 29988. The same test sets were used between models for each run. The bold numbers indicate the model with the highest sensitivity for each category of test data. One run of a baseline model trained on male and female data is added for comparison only.}
\end{table*}

The results of this can be seen in Table~\ref{toxicity-sensitivity-table}, where we observe similar results to \citet{binns2017like}, showing that models consistently perform worse on female-annotated test data compared to male-annotated test data. This could be due to the greater diversity of opinions in female-annotated data resulting from low inter-annotator agreement \citep{binns2017like}, in addition to the ability of the model to associate offensive words with male annotations making it easier to classify toxic comments annotated by men. We also note that female-annotated models perform 1.8\% $\pm$ 0.6\% better on average, suggesting they are less dependent on the presence of offensive words in test data for classification.

We observe that when the offensive words in the training and test data are removed, the toxic comments without offensive words become more difficult to correctly classify than those with offensive words. We also find that models trained on data without offensive words have a 0.4\% higher sensitivity on average on unmodified test data than the equivalent model trained on data with offensive words. The performance of BERTMaleNoProfanity surpasses the performance of BERTMale on every set. BERTFemaleNoProfanity has a similar performance on the unmodified data as BERTFemale, despite the lack of offensive words in the training data. BERTFemaleNoProfanity outperforms BERTFemale by 0.1272 and 0.1162 on the modified male and female test data respectively. This is due to the model relying on factors other than the offensive words for toxicity classification. 

\section{Discussion}

Toxic language detection is a highly subjective task, with majority opinions and levels of agreement varying within and between demographic groups. We highlight this by analysing the annotations of different genders in the chosen corpus, noting that the number of female annotators is outweighed by the number of male annotators, and that the female annotators are more likely to label a comment as toxic than their male counterparts. This information could be leveraged by moderation systems by taking into account the demographic group the reader of a comment belongs to before determining the toxicity threshold at which a comment is removed from the system.

Our findings indicate that the BERT-based model associates comments that contain offensive words with male annotators, despite the data showing that both male and female annotators label comments containing high numbers of offensive words as toxic. We demonstrate that the most offensive words are attributed to male annotators, which causes the model to output skewed predictions indicating that most comments have been annotated by men despite the training data being balanced between both genders.

We note that the male annotators in this corpus display a greater level of inter-annotator agreement than the female annotators which may contribute to the tendency of the model to predict the gender of an annotator as male. This bias indicates that toxicity models trained on this corpus will be more influenced by the opinions of male annotators, as the diversity of views given by the female annotators makes them unlikely to hold the majority opinion, and those who label comments containing offensive words as toxic are perceived to be male by the model.

We find that removing the offensive words from the training data produces a model that demonstrates less bias overall than the original model but exhibits the most variation in the results of any of the implemented models. We find that removing the most toxic data in addition to removing the offensive words in the training data produces the model with the least bias, showing that comments containing high numbers of offensive words are far less attributed to male annotators than in the original model.

Applying the discovered associations between gender and offensive language to models tasked with classifying the toxicity of comments, we find that toxic comments annotated by men are easier to classify than those annotated by women. Conversely, we find that models trained exclusively on female-annotated data display a better performance than models trained entirely on male-annotated data. This is in part due to the associations between male annotators and offensive language distracting the model from other aspects of toxic comments.

Finally, we show that while it is harder to correctly classify toxic data after the removal of offensive words, models trained on this data show a comparable performance to models trained on unmodified data. Combining these results with those of the gender predicting models, we see that removing offensive words from the training data of a model is an effective way of reducing the bias towards the opinions of male annotators without compromising the performance of the model on toxic data.

We note that this approach does not remove all bias in the model, for example we did not address the male bias present in the model due to the contextual relationships between words found in the training data \citep{kurita-etal-2019-measuring}. However, this paper provides an insight into the gender associations that can be present in a model and the methods that can be used to investigate and minimise bias in any classification system reliant on annotators.

We recommend that the demographics of the annotators be collected and reported as part of labelled datasets. This is particularly relevant in problems which rely on the subjective opinion of the annotator like toxic language detection. 

\section{Conclusion}

In this paper we seek to quantify the gender bias in toxic language detection systems present as a result of differences in the opinions held by distinct demographic groups of annotators in the corpus and aim to minimise this bias without compromising the performance of the model. We identify differences between the annotation styles of men and women in the chosen corpus and determine that this causes a bias towards the opinions of men. We discover associations between the male bias and the use of offensive language in toxic comments, applying this knowledge to a toxic language classifier to demonstrate an effective way to reduce gender bias without compromising the performance of the model.

Future work on annotator bias should examine other demographic variables present in the pool of annotators such as race, age or level of education and analyse the extent to which certain groups may be excluded or have their opinions overlooked by the model. This could be extended by researching the connection between the demographic identities of annotators and the identities referenced in comments to see where prejudice occurs. Those implementing toxic language detection systems would be advised to consider the types of bias present in their model and personalise moderation based on the identities of those authoring or viewing comments.



\bibliographystyle{acl_natbib}
\bibliography{final-submission}

\begin{thebibliography}{22}
\expandafter\ifx\csname natexlab\endcsname\relax\def\natexlab#1{#1}\fi

\bibitem[{Aroyo and Welty(2013)}]{aroyo2013crowd}
Lora Aroyo and Chris Welty. 2013.
\newblock Crowd truth: Harnessing disagreement in crowdsourcing a relation
  extraction gold standard.
\newblock In \emph{Proceedings of ACM Web Science 2013 Conference}.

\bibitem[{Balayn et~al.(2018)Balayn, Mavridis, Bozzon, Timmermans, and
  Szl{\'a}vik}]{balayn2018characterising}
Agathe Balayn, Panagiotis Mavridis, Alessandro Bozzon, Benjamin Timmermans, and
  Zolt{\'a}n Szl{\'a}vik. 2018.
\newblock Characterising and mitigating aggregation-bias in crowdsourced
  toxicity annotations.
\newblock In \emph{Proceedings of the 1st Workshop on Subjectivity, Ambiguity
  and Disagreement in Crowdsourcing, and Short Paper Proceedings of the 1st
  Workshop on Disentangling the Relation Between Crowdsourcing and Bias
  Management}, volume 2276. CEUR.

\bibitem[{Binns et~al.(2017)Binns, Veale, Van~Kleek, and
  Shadbolt}]{binns2017like}
Reuben Binns, Michael Veale, Max Van~Kleek, and Nigel Shadbolt. 2017.
\newblock Like trainer, like bot? inheritance of bias in algorithmic content
  moderation.
\newblock In \emph{International conference on social informatics}, pages
  405--415. Springer.

\bibitem[{Blodgett et~al.(2020)Blodgett, Barocas, Daum{\'e}~III, and
  Wallach}]{blodgett-etal-2020-language}
Su~Lin Blodgett, Solon Barocas, Hal Daum{\'e}~III, and Hanna Wallach. 2020.
\newblock \href {https://doi.org/10.18653/v1/2020.acl-main.485} {Language
  (technology) is power: A critical survey of {``}bias{''} in {NLP}}.
\newblock In \emph{Proceedings of the 58th Annual Meeting of the Association
  for Computational Linguistics}, pages 5454--5476, Online. Association for
  Computational Linguistics.

\bibitem[{Bolukbasi et~al.(2016)Bolukbasi, Chang, Zou, Saligrama, and
  Kalai}]{bolukbasi2016man}
Tolga Bolukbasi, Kai-Wei Chang, James Zou, Venkatesh Saligrama, and Adam Kalai.
  2016.
\newblock Man is to computer programmer as woman is to homemaker? debiasing
  word embeddings.
\newblock In \emph{Proceedings of the 30th International Conference on Neural
  Information Processing Systems}, NIPS'16, page 4356–4364, Red Hook, NY,
  USA. Curran Associates Inc.

\bibitem[{Devlin et~al.(2019)Devlin, Chang, Lee, and
  Toutanova}]{devlin-etal-2019-bert}
Jacob Devlin, Ming-Wei Chang, Kenton Lee, and Kristina Toutanova. 2019.
\newblock \href {https://doi.org/10.18653/v1/N19-1423} {{BERT}: Pre-training of
  deep bidirectional transformers for language understanding}.
\newblock In \emph{Proceedings of the 2019 Conference of the North {A}merican
  Chapter of the Association for Computational Linguistics: Human Language
  Technologies, Volume 1 (Long and Short Papers)}, pages 4171--4186,
  Minneapolis, Minnesota. Association for Computational Linguistics.

\bibitem[{Dixon et~al.(2018)Dixon, Li, Sorensen, Thain, and
  Vasserman}]{dixon2018measuring}
Lucas Dixon, John Li, Jeffrey Sorensen, Nithum Thain, and Lucy Vasserman. 2018.
\newblock Measuring and mitigating unintended bias in text classification.
\newblock In \emph{Proceedings of the 2018 AAAI/ACM Conference on AI, Ethics,
  and Society}, pages 67--73.

\bibitem[{Kolhatkar et~al.(2020)Kolhatkar, Thain, Sorensen, Dixon, and
  Taboada}]{kolhatkar2020classifying}
Varada Kolhatkar, Nithum Thain, Jeffrey Sorensen, Lucas Dixon, and Maite
  Taboada. 2020.
\newblock Classifying constructive comments.
\newblock \emph{arXiv preprint arXiv:2004.05476}.
\newblock Unpublished.

\bibitem[{Kurita et~al.(2019)Kurita, Vyas, Pareek, Black, and
  Tsvetkov}]{kurita-etal-2019-measuring}
Keita Kurita, Nidhi Vyas, Ayush Pareek, Alan~W Black, and Yulia Tsvetkov. 2019.
\newblock \href {https://doi.org/10.18653/v1/W19-3823} {Measuring bias in
  contextualized word representations}.
\newblock In \emph{Proceedings of the First Workshop on Gender Bias in Natural
  Language Processing}, pages 166--172, Florence, Italy. Association for
  Computational Linguistics.

\bibitem[{Manzini et~al.(2019)Manzini, Yao~Chong, Black, and
  Tsvetkov}]{manzini-etal-2019-black}
Thomas Manzini, Lim Yao~Chong, Alan~W Black, and Yulia Tsvetkov. 2019.
\newblock \href {https://doi.org/10.18653/v1/N19-1062} {Black is to criminal as
  caucasian is to police: Detecting and removing multiclass bias in word
  embeddings}.
\newblock In \emph{Proceedings of the 2019 Conference of the North {A}merican
  Chapter of the Association for Computational Linguistics: Human Language
  Technologies, Volume 1 (Long and Short Papers)}, pages 615--621, Minneapolis,
  Minnesota. Association for Computational Linguistics.

\bibitem[{Mishra et~al.(2018)Mishra, Yannakoudakis, and
  Shutova}]{mishra-etal-2018-neural}
Pushkar Mishra, Helen Yannakoudakis, and Ekaterina Shutova. 2018.
\newblock \href {https://doi.org/10.18653/v1/W18-5101} {Neural character-based
  composition models for abuse detection}.
\newblock In \emph{Proceedings of the 2nd Workshop on Abusive Language Online
  ({ALW}2)}, pages 1--10, Brussels, Belgium. Association for Computational
  Linguistics.

\bibitem[{Nobata et~al.(2016)Nobata, Tetreault, Thomas, Mehdad, and
  Chang}]{nobata2016abusive}
Chikashi Nobata, Joel Tetreault, Achint Thomas, Yashar Mehdad, and Yi~Chang.
  2016.
\newblock Abusive language detection in online user content.
\newblock In \emph{Proceedings of the 25th international conference on world
  wide web}, pages 145--153.

\bibitem[{Pavlopoulos et~al.(2017)Pavlopoulos, Malakasiotis, and
  Androutsopoulos}]{pavlopoulos-etal-2017-deep}
John Pavlopoulos, Prodromos Malakasiotis, and Ion Androutsopoulos. 2017.
\newblock \href {https://doi.org/10.18653/v1/W17-3004} {Deep learning for user
  comment moderation}.
\newblock In \emph{Proceedings of the First Workshop on Abusive Language
  Online}, pages 25--35, Vancouver, BC, Canada. Association for Computational
  Linguistics.

\bibitem[{Pavlopoulos et~al.(2020)Pavlopoulos, Sorensen, Dixon, Thain, and
  Androutsopoulos}]{pavlopoulos-etal-2020-toxicity}
John Pavlopoulos, Jeffrey Sorensen, Lucas Dixon, Nithum Thain, and Ion
  Androutsopoulos. 2020.
\newblock \href {https://doi.org/10.18653/v1/2020.acl-main.396} {Toxicity
  detection: Does context really matter?}
\newblock In \emph{Proceedings of the 58th Annual Meeting of the Association
  for Computational Linguistics}, pages 4296--4305, Online. Association for
  Computational Linguistics.

\bibitem[{Sap et~al.(2019)Sap, Card, Gabriel, Choi, and
  Smith}]{sap-etal-2019-risk}
Maarten Sap, Dallas Card, Saadia Gabriel, Yejin Choi, and Noah~A. Smith. 2019.
\newblock \href {https://doi.org/10.18653/v1/P19-1163} {The risk of racial bias
  in hate speech detection}.
\newblock In \emph{Proceedings of the 57th Annual Meeting of the Association
  for Computational Linguistics}, pages 1668--1678, Florence, Italy.
  Association for Computational Linguistics.

\bibitem[{Schmidt and Wiegand(2017)}]{schmidt2017survey}
Anna Schmidt and Michael Wiegand. 2017.
\newblock A survey on hate speech detection using natural language processing.
\newblock In \emph{Proceedings of the fifth international workshop on natural
  language processing for social media}, pages 1--10.

\bibitem[{Sun et~al.(2019)Sun, Gaut, Tang, Huang, ElSherief, Zhao, Mirza,
  Belding, Chang, and Wang}]{sun-etal-2019-mitigating}
Tony Sun, Andrew Gaut, Shirlyn Tang, Yuxin Huang, Mai ElSherief, Jieyu Zhao,
  Diba Mirza, Elizabeth Belding, Kai-Wei Chang, and William~Yang Wang. 2019.
\newblock \href {https://doi.org/10.18653/v1/P19-1159} {Mitigating gender bias
  in natural language processing: Literature review}.
\newblock In \emph{Proceedings of the 57th Annual Meeting of the Association
  for Computational Linguistics}, pages 1630--1640, Florence, Italy.
  Association for Computational Linguistics.

\bibitem[{Sundararajan et~al.(2017)Sundararajan, Taly, and
  Yan}]{sundararajan2017axiomatic}
Mukund Sundararajan, Ankur Taly, and Qiqi Yan. 2017.
\newblock Axiomatic attribution for deep networks.
\newblock In \emph{International Conference on Machine Learning}, pages
  3319--3328. PMLR.

\bibitem[{Wolf et~al.(2020)Wolf, Debut, Sanh, Chaumond, Delangue, Moi, Cistac,
  Rault, Louf, Funtowicz, Davison, Shleifer, von Platen, Ma, Jernite, Plu, Xu,
  Le~Scao, Gugger, Drame, Lhoest, and Rush}]{wolf-etal-2020-transformers}
Thomas Wolf, Lysandre Debut, Victor Sanh, Julien Chaumond, Clement Delangue,
  Anthony Moi, Pierric Cistac, Tim Rault, Remi Louf, Morgan Funtowicz, Joe
  Davison, Sam Shleifer, Patrick von Platen, Clara Ma, Yacine Jernite, Julien
  Plu, Canwen Xu, Teven Le~Scao, Sylvain Gugger, Mariama Drame, Quentin Lhoest,
  and Alexander Rush. 2020.
\newblock \href {https://doi.org/10.18653/v1/2020.emnlp-demos.6} {Transformers:
  State-of-the-art natural language processing}.
\newblock In \emph{Proceedings of the 2020 Conference on Empirical Methods in
  Natural Language Processing: System Demonstrations}, pages 38--45, Online.
  Association for Computational Linguistics.

\bibitem[{Wulczyn et~al.(2017)Wulczyn, Thain, and Dixon}]{wulczyn2017ex}
Ellery Wulczyn, Nithum Thain, and Lucas Dixon. 2017.
\newblock Ex machina: Personal attacks seen at scale.
\newblock In \emph{Proceedings of the 26th international conference on world
  wide web}, pages 1391--1399.

\bibitem[{Zhao et~al.(2018)Zhao, Wang, Yatskar, Ordonez, and
  Chang}]{zhao-etal-2018-gender}
Jieyu Zhao, Tianlu Wang, Mark Yatskar, Vicente Ordonez, and Kai-Wei Chang.
  2018.
\newblock \href {https://doi.org/10.18653/v1/N18-2003} {Gender bias in
  coreference resolution: Evaluation and debiasing methods}.
\newblock In \emph{Proceedings of the 2018 Conference of the North {A}merican
  Chapter of the Association for Computational Linguistics: Human Language
  Technologies, Volume 2 (Short Papers)}, pages 15--20, New Orleans, Louisiana.
  Association for Computational Linguistics.

\bibitem[{Zorian and Bikkanur(2019)}]{zorian2019debiasing}
Apik~Ashod Zorian and Chandra~Shekar Bikkanur. 2019.
\newblock Debiasing personal identities in toxicity classification.
\newblock \emph{arXiv preprint arXiv:1908.05757}.
\newblock Unpublished.

\end{thebibliography}


\end{document}